\documentclass[sn-mathphys,Numbered,pdflatex]{sn-jnl}

\usepackage{graphicx}%
\usepackage{multirow}%
\usepackage{amsmath,amssymb,amsfonts}%
\usepackage{amsthm}%
\usepackage{mathrsfs}%
\usepackage[title]{appendix}%
\usepackage{xcolor}%
\usepackage{textcomp}%
\usepackage{manyfoot}%
\usepackage{booktabs}%
\usepackage{algorithm}%
\usepackage{algorithmicx}%
\usepackage{algpseudocode}%
\usepackage{listings}%
\usepackage{subfigure}
\usepackage{multirow}
\usepackage[normalem]{ulem}
\useunder{\uline}{\ul}{}
\usepackage{pdflscape}%
\usepackage{fancyref}%
\usepackage[format=hang,font=small,labelfont=bf,labelsep=space]{caption}%
\usepackage{float}%
\usepackage{adjustbox}%
\theoremstyle{plain}
\theoremstyle{thmstyleone}%
\theoremstyle{thmstyletwo}%

\theoremstyle{thmstylethree}%

\begin{document}

\title[Article Title]{AI Powered Road Network Prediction with Multi-Modal Data}


\author*[1,3]{\fnm{Necip Enes} \sur{Gengec}}\email{gengec@itu.edu.tr}

\author[2]{\fnm{Ergin} \sur{Tari}}\email{tari@itu.edu.tr}

\author[3]{\fnm{Ulas} \sur{Bagci}}\email{ulas.bagci@northwestern.edu}

\affil*[1]{\orgdiv{Graduate School of Engineering and Technology}, \orgname{Istanbul Technical University}, \orgaddress{\city{Istanbul}, \country{Turkey}}}

\affil[2]{\orgdiv{Department of Geomatics Engineering}, \orgname{Istanbul Technical University}, \orgaddress{\city{Istanbul}, \country{Turkey}}}

\affil[3]{\orgdiv{Machine and Hybrid Intelligence Lab}, \orgname{Northwestern University}, \orgaddress{\city{Chicago}, \state{Illinois}, \country{USA}}}


\abstract{This study presents an innovative approach for automatic road detection with deep learning, by employing fusion strategies for utilizing both lower-resolution satellite imagery and GPS trajectory data, a concept never explored before. We rigorously investigate both early and late fusion strategies, and assess deep learning based road detection performance using different fusion settings. Our extensive ablation studies assess the efficacy of our framework under diverse model architectures, loss functions, and geographic domains (Istanbul and Montreal). For an unbiased and complete evaluation of road detection results, we use both region-based and boundary-based evaluation metrics for road segmentation. The outcomes reveal that the ResUnet model outperforms U-Net and D-Linknet in road extraction tasks, achieving superior results over the benchmark study using low-resolution Sentinel-2 data. This research not only contributes to the field of automatic road detection but also offers novel insights into the utilization of data fusion methods in diverse applications.}




\maketitle

\section{Introduction}

Digital maps are used in wide range of applications including navigation, urban planning, disaster management and response, and many more with "road network data" serving as a primary component of these maps \cite{survey_wiley, d_linknet, deepglobe}. Road network data can be produced manually through digitization or field surveys, crowd-sourced, or automatically detected through aerial/satellite imagery and/or using GPS trajectories.

While its significance is bold, analyzing road network data can be quite challenging with manual efforts. Hence, automatic detection of road networks from images has recently been adopted due to its cost efficiency. The success of emerging artificial intelligence (AI)/deep learning (DL) methods has played a primary role in this switch \cite{Liu2022}. In these applications, the first and the major step is to segment (delineate) satellite or aerial images using supervised deep learning models. High-resolution satellite imagery is more often used and desired in such applications than other imaging modalities \cite{mnih, deepglobe, resunet, sun2018} but automated methods with high-resolution satellite imagery is still costly. Because of high cost of such images, using lower-resolution imagery such as freely available Sentinel-2 \cite{Ayala2021_sentinel, Johnson2022_sentinel} becomes an attractive research area with a few existing studies. The use of low-resolution images presents certain challenges including having lower resolution for details, thus low accuracy in quantitative measures coupled to it. However, it provides also potential opportunities to research on. For example, Sentinel-2 is freely available and can provide broad coverage (more global). Further, Sentinel-2 provides a better temporal resolution and it is multi-spectral in nature (i.e., capturing several spectral bands). Practical and cost-effective nature of the low-resolution imagery is opening new and unexplored doors for research community. That being said, current efforts in this domain and particularly in road detection tasks is limited and in early steps; further research on improving the automatic road detection task with lower-resolution data can provide more cost-effective solutions. In this paper, our effort is within this research line: we aim to develop cost-effective AI solution for road network prediction with multi-modal data.

\begin{figure}[H]
    \centering
    \includegraphics[width=0.9\textwidth]{ 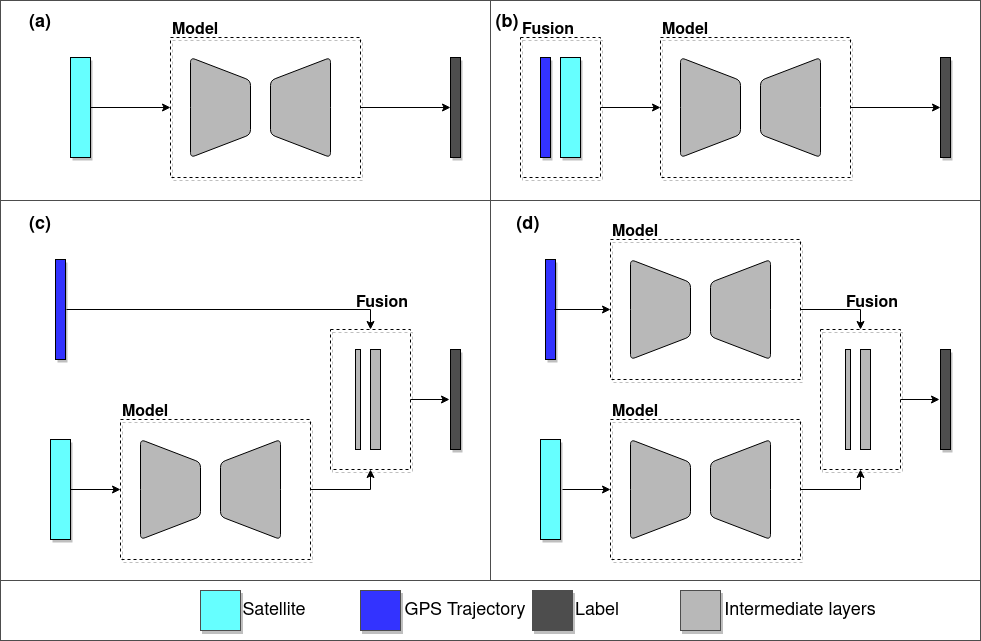}
    \caption{(a) Baseline model with satellite image input, (b) baseline model with early fusion of satellite image and GPS trajectory input, (c) late fusion Type-1 (model applied only to satellite image stream) with satellite image and GPS trajectory input, (d) late fusion Type-2 (model applied to both satellite image and GPS trajectory streams) with satellite image and GPS trajectory input.} \label{fig:main_data_flow}
\end{figure}

"GPS trajectory data" is another source used in road network segmentation. Different methods are used to detect roads, including point clustering \cite{gps_only_2}, kernel density estimation (KDE) \cite{gps_only_4}, graph-based road generation \cite{gps_only_3}, and point matching \cite{gps_only_1}. In addition, deep learning methods are used for road segmentation over rasterized GPS trajectory data fusion with satellite imagery \cite{sun2019}. High-resolution satellite imagery is still an expensive choice in these cases. To our knowledge, no study has been conducted yet using lower-resolution satellite imagery and GPS trajectory fusion for automatic road detection. In different fields, fusion operations are commonly used in ad-hoc manner. For example, early fusion (Figure \ref{fig:main_data_flow}b) is the prominent method in combining varying data sources. However, optimal fusion strategy is often unknown especially when data sources have some common overlaps. In other words, the effect of fusion at later stages and success of alternative fusion operations on segmentation is largely unknown. We speculate that exploring such gaps may improve the road detection task.

\textbf{The overall goal }of this study is therefore to introduce an innovative approach for automatic road detection and segmentation by fusing lower-resolution satellite imagery with GPS trajectory data, an area yet unexplored in the current landscape of studies. We will investigate both early and late fusion strategies for low resolution satellite imagery with GPS trajectory data (Figure \ref{fig:main_data_flow}c and \ref{fig:main_data_flow}d) and explore road segmentation performance in depth using relatively lower-resolution satellite imagery in different fusion settings. In our ablation studies, the efficacy of this framework is tested under various settings of model architectures and loss functions in different geographic domains.

\section{Related Work}

The automatic detection of road networks has become an increasingly popular research topic due to its practical applications \cite{Liu2022, survey_wiley}. In recent years, many studies have focused on the use of deep learning methods to extract road networks from various data sources such as satellite imagery and GPS trajectory data \cite{sun2019, gps_sat_fusion1, gps_sat_fusion2, gps_sat_fusion3}. Even with deep learning, advanced artificial intelligence methodologies solving complex problems at scale with highly accurate manner, the problem is still solved at sub-optimally pace because of highly variable image qualities across different data sources, even within the data source, complexity of the road features, problems such as occlusion, lighting differences, and other similar-looking features. To this end, existing studies presented several fusion methods to integrate GPS trajectory data into satellite imagery to improve the accuracy of road extraction. Some studies focused on exploring different loss functions that might be more suitable for the road extraction tasks. Last, but not least, it is worth to revisit the evaluation metrics for deep learning based segmentation strategies as they are crucial for measuring the effectiveness of different road extraction methods. In this section, we provide a review of the relevant literature in all these areas.

\subsection{Road extraction using satellite imagery with deep learning}

Mnih and Hinton's pioneering work (\citeyear{mnih}) was the first significant study to apply deep neural networks to road extraction from satellite imagery. Since then, many studies have used different deep network architectures to improve the performance of road extraction. \textit{U-Net}, the widely used U-shaped deep network, was originally developed for medical image segmentation \cite{unet}. Later, it has been applied to road segmentation in different studies. Literature becomes increasingly vast in methods that relying on U-Net. For instance, Residual \textit{U-Net} (\textit{ResUnet}) has been used in road segmentation with satellite imagery which is one variant of \textit{U-Net} that uses residual units to enhance segmentation results \cite{resunet}. \textit{D-Linknet} is another U-shaped network that uses dilated convolutions and has been frequently used as a benchmark model \cite{d_linknet}. Other notable studies are \textit{BiHRnet} \cite{bi-hrnet}, \textit{HsgNet} \cite{hsgnet}, \textit{RADANet} \cite{sat_only3}, \textit{SDUNet} \cite{sat_only5}, and the study of \cite{sat_only2}. In more recent years, we are witnessing a huge swamp from CNN based architectures to Transformers based architectures due to their self-attention mechanisms and better performances when the architectures are not-so-deep. Despite their success, there is a high computational burden in Transformers as well as more data requirement, not allowing them to be easily adapted for multidimensional data. \textit{BDTNet} \cite{sat_only1}, \textit{RoadFormer} \cite{sat_only4}, and \textit{Seg-Road} \cite{sat_only6} are recent examples of the latest transformer-based models that have been applied to road segmentation.

In the context of image resolution, deep learning-based approaches are already applied to high-resolution satellite imagery \cite{mnih,resunet,bi-hrnet,hsgnet,sat_only3,sat_only5,sat_only2,sat_only1,sat_only4}. There are only a few approaches that use lower resolution satellite imagery such as \cite{Ayala2021_sentinel}, \cite{Ayala2021_sentinel_2} and \cite{Johnson2022_sentinel}. These examples use Sentinel-2 data for road extraction as an input to either \textit{U-Net} or \textit{HRNet}.

\subsection{Road extraction using satellite imagery and GPS Trajectory with deep learning}
Deep learning architectural engineering becomes a de facto strategy for improved road segmentation performance \cite{Liu2022}. For instance, \cite{sun2019} proposed a U-shaped architecture with 1D convolution, where satellite imagery and GPS trajectory data are fed into the network as concatenated image layers. \cite{gps_sat1} used a similar approach, utilizing a \textit{U-Net} model with refined labels. \textit{D-Linknet}, which incorporates concatenated satellite imagery and GPS trajectory data, has been frequently used and extended in recent studies such as \textit{FuNet} \cite{gps_sat2}, \textit{RING-Net} \cite{gps_sat3}, and \cite{gps_sat4}. Other studies have proposed novel techniques to incorporate GPS trajectory data, such as \cite{gps_sat_fusion1}, \cite{gps_sat_fusion2}, and \cite{gps_sat_fusion3}. These studies demonstrate the potential benefits of combining GPS trajectory data with satellite imagery for improved road segmentation accuracy. Despite their benefits, none of these studies have reported using lower resolution satellite imagery in conjunction with GPS trajectory data. Also, 1D convolution is more appropriate for GPS trajectory data while not for imaging data indicating potential sub-optimality in fusing the data.

\subsection{Multi modal data fusion}
Multi-modal data can be fused within deep learning models. Theoretically, the fusion process can occur at various stages within the model, employing different fusion methods. These fusion stages can be categorized as early, late and hybrid fusion \cite{fusion_types}. In early fusion, the fusion takes place at the beginning of the model (Figure \ref{fig:main_data_flow}b) where as late fusion occurs at the end, just before the output layer (Figure \ref{fig:main_data_flow}c and \ref{fig:main_data_flow}d). Hybrid fusion involves a more complex flow and can be summarized as fusion that takes place at the intermediate stages of the model. When considering fusion methods, multiple matrix operations can be utilized based on the desired outcome, often involving a trial and error. Concatenation is the most frequently preferred fusion method \cite{sun2019, gps_sat_fusion1, gps_sat_fusion2, gps_sat_fusion3}.

In the context of road extraction using satellite imagery and GPS trajectory, different stages of fusion methods have been tested. \cite{gps_sat_fusion1} proposed their own method and evaluated its accuracy in comparison to early and late fusion alternatives, utilizing concatenation as the fusion method. Their study found that early fusion provided slightly better IoU results when compared to late fusion. In another study, \cite{gps_sat_fusion2} examined early and late fusion in their \textit{DeepDualMapper} study. They employed concatenation for early fusion and averaging for late fusion as the fusion method. Similar to \cite{gps_sat_fusion1}, \cite{gps_sat_fusion2} achieved the superior results with early fusion. Furthermore, \cite{gps_sat_fusion3} explored early, deep, and vanilla fusion in their study. Deep fusion represents an example of hybrid fusion while vanilla fusion is a late fusion variant that employs intersection as the fusion method. In this study, early fusion outperformed vanilla fusion in terms of recall and $F_{1}$. Literature shows that research on fusion stages is limited. As a note, the fusion methods utilized in these studies mostly revolve around concatenation only.

\subsection{Loss functions}
Loss functions are needed in the optimization of deep neural networks \cite{goodfellow_dl}. Numerous loss functions have been proposed according to the specific task at hand. In the context of road extraction, mean square error (MSE) \cite{resunet} and binary cross-entropy (BCE) \cite{survey_wiley} are two commonly used functions. BCE is generally regularized with an additional loss function such as Dice \cite{sun2018, d_linknet, hsgnet} or $L_{2}$ norm \cite{bi-hrnet}. \cite{deepglobe} employed a focal loss function, a BCE variant that addresses class imbalance issues. Furthermore, researchers have proposed application-specific tailored loss functions by combining multiple loss functions \cite{gps_sat_fusion3, gps_sat3, gps_sat1} when necessary. To our best of knowledge, no study has been conducted to comprehensively evaluate their performance in road extraction using deep learning. Our study fills this research gap.

\subsection{Evaluation metrics}
Evaluation metrics are essential for monitoring the performance of a given model. Various metrics have been adopted in segmentation tasks \cite{cnn_accuracy1, cnn_accuracy2} in general. Precision and recall are considered as fundamental metrics in many road extraction studies. These metrics are often employed alongside additional metrics such as the $F_{1}$ score and/or intersection over union (IoU) \cite{gps_sat_fusion3, hsgnet, bi-hrnet} in practice. Precision and recall are used to calculate the $F_{1}$ score, which is calculated by the harmonic mean of these two metrics. IoU represents the ratio between the intersection and the union of the ground truth and predicted segments. In some studies, IoU is used as the sole metric \cite{d_linknet, sun2019, sun2018}. Occasionally, custom metrics are employed, such as the break-even point/relaxed precision \cite{resunet}, global IoU \cite{gps_sat_fusion1} or average path length similarity (APLS) \cite{spacenet_1, spacenet_2, bi-hrnet}. However, the adoption of these metrics remains limited.

\cite{metrics_reloaded} developed a framework to guide the selection of appropriate metrics for different machine learning tasks. For segmentation tasks, the framework suggests using an region-based metric such as IoU or $F_{1}$ score, complemented by a boundary-based metric. The inclusion of a boundary-based metric helps to address the issues caused by the lack of shape awareness in region-based metrics. Notably, there are no literature examples of boundary-based metrics being used in the context of road extraction. In order to complement full evaluation spectrum, this metric and region-based metric are comprehended in our study.

\subsection{Benchmark dataset}
Benchmarking serves the purpose of facilitating fair comparisons and validations among different models under the same conditions, thereby enabling the identification of strengths and weaknesses. Several benchmark dataset are available for road extraction from satellite imagery \cite{Liu2022}. Massachusetts \cite{mnih}, DeepGlobe \cite{deepglobe} and SpaceNet \cite{spacenet_2} dataset are the leading examples which are widely used as benchmark. These dataset comprise high-resolution satellite imagery. In couple couple of research satellite imagery extracted from Google Maps API from different zoom levels is used in road extraction. \cite{ozturk2022_istanbul} used the Google Maps API to obtain satellite imagery for the road extraction task in Istanbul, while \cite{gps_sat_fusion2} acquired data from Porto, Shanghai and Singapore in the same method and they conducted their study with additional GPS trajectory data. \cite{Ayala2021_sentinel, Ayala2021_sentinel_2} conducted road extraction study using low-resolution Sentinel-2 data.

The benchmark dataset available in the literature are predominantly based on high-resolution imagery, which can be costly to acquire for real-world applications. Approaches such as those employed by \cite{ozturk2022_istanbul} and \cite{gps_sat_fusion2} are not viable for all scenarios. The utilization of freely available low-resolution Sentinel-2 data or similar low-resolution satellite imagery sources is noteworthy, although the availability of GPS trajectory data is essential to support research in the multi-modal domain. Furthermore, such dataset should cover multiple geographies to enhance studies that measure the generalizability of model performance across different dataset. Our study provides a benchmark dataset which consist of low-resolution satellite imagery and GPS trajectory data from two different locations which is filling the gap in literature. 

\subsection{Our contributions}

The main novelty of our study lies in its innovative use of lower-resolution satellite imagery and GPS trajectory fusion for road detection and quantification via segmentation. In the light of relevant studies and their limitations, our study has the following major contributions:
\begin{enumerate}
    \item We extensively investigate the impact of GPS trajectory data on road extraction using low-resolution satellite imagery (Sentinel-2). Through this, we anticipate to initiate a new wave of studies focused on exploiting lower-resolution image and GPS trajectory data, ultimately contributing broader advance of automatic road detection methods.
    \item We carefully design fusion architectures (early fusion, late fusion Type-1/2) consisting of the state-of-the-art architectures (\textit{U-Net}, \textit{ResUnet}, \textit{D-Linknet}) with various loss functions (MSE, BCE, Focal loss) using both Sentinel-2 data and GPS trajectory data. The fusion architectures is expected to amplify the efficacy of road detection.
    \item We assess the fusion performance of the ablation models by employing fusion techniques at different stages and utilizing various fusion methods (e.g., early fusion, late fusion) as illustrated in Figure~\ref{fig:main_data_flow}.
    \item We provide a novel benchmark dataset and test the generalization ability of the models on a newly developed benchmark dataset that incorporates multi-modal data, including GPS trajectory and Sentinel-2 data from diverse geographic locations (Istanbul and Montreal).
    \item Due to inherent limitations of traditional evaluation metrics for segmentation tasks, we postulate a full spectrum segmentation evaluation strategy by using both region and boundary-based metrics, giving broader understanding of segmentation methods under various conditions. We propose to use both region (IoU), and boundary based methods (Boundary-IoU) together to give a better understanding and fair evaluation of methods.
\end{enumerate}

By addressing these objectives, our study aims to (1) explore the influence of GPS trajectory data by (2) evaluating different deep learning architectures, (3) comparing loss functions, (4) analyzing fusion techniques, and their generalization capabilities, and (5) applying a new type of evaluation metric in the road extraction research.

\section{Methodology}
\label{sec:methodology}
In this section, we delve into the methodology employed to accomplish the research objectives of this study.

\subsection{Choosing segmentation models}

Based on the state of the art algorithms, we employed \textit{U-Net}, \textit{ResUnet} and \textit{D-Linknet} in order to assess their strengths and weaknesses in the road extraction using satellite imagery and GPS trajectory data. Briefly, these methods are described as follows.

\textbf{\textit{U-Net} }is a convolutional neural network architecture that incorporates both convolutional and up-convolutional layers, connected by skip connections \cite{unet}. It consists of an encoder, a bottleneck, a decoder, and skip connections between the encoder and decoder parts. Although initially developed for biomedical image segmentation, \textit{U-Net} has been successfully applied in various domains. The original \textit{U-Net} is trained on the RGB data, where each color layer is stacked into a 3D tensor, yielding binary predictions.

\textbf{\textit{ResUnet}} is a variant of \textit{U-Net} specifically designed for road extraction from satellite imagery. \textit{ResUnet} improves upon \textit{U-Net} by incorporating residual units \cite{resunet}. Residual learning or residual unit, first introduced by \cite{resnet}, addresses the problem of overfitting in large deep neural networks. In \textit{ResUnet}, the plain neural units in \textit{U-Net} are replaced with identity-mapped replicas of the same units, known as residual units \cite{resunet}. This addition leads to significant improvement in IoU.

\textbf{\textit{D-Linknet}} is another U-shaped segmentation model developed for road extraction, building upon the success of its predecessor, \textit{Linknet} \cite{d_linknet}. \textit{D-Linknet} introduces a dilated convolution convolution block in the bottleneck of \textit{U-Net} along with the residual units. Additionally, \textit{D-Linknet} leverages transfer learning, where the encoder part of the model is initialized with a ResNet34 pretrained on the ImageNet dataset. \textit{D-Linknet} achieved the best results in the DeepGlobe Road Extraction Challenge - 2018 \cite{deepglobe}, and subsequent improvements have been made by other researchers \cite{hsgnet, bi-hrnet}.

\subsection{Loss functions details}
\label{sec:method_loss_functions}

Mean square error (MSE), binary cross-entropy (BCE), and focal loss were utilized in this study to train the networks and assess their performance in road extraction tasks.

\textbf{MSE}, which is an example of mean bias error (MBE) losses \cite{loss_function_tax_survey}, is calculated as the sum of squared errors between predictions and the ground truth. It can be defined by the following equation \cite{resunet}:
\newline

\begin{equation}
    L_{MSE}(W)=\frac{1}{N}\sum_{i=1}^{N}\|Net(I_{i};W)-s_{i}\|^2,
\end{equation}

\noindent where $Net(I_{i};W)$ represents the segmentation, $s_{i}$ denotes the ground truth, and $N$ is the number of training examples.

\textbf{BCE} is a probabilistic loss function \cite{loss_function_tax_survey} used to measure the difference between two probability distributions \cite{loss_function_survey}. It is defined as:
\newline

\begin{equation}
    L_{BCE}(y,\hat{y})=-(y\log{(\hat{y})}+(1-y)\log{(1-\hat{y})}),
\end{equation}

\noindent where $y$ represents the ground truth and $\hat{y}$ represents the predictions.

In the context of segmentation tasks, the available classes in the data are often imbalanced. For example, in the road extraction, foreground pixels are more frequent when compared to road pixels. \textbf{Focal loss} is a loss function designed to address such class imbalance \cite{focal_loss}:
\newline

\begin{equation}
    L_{FL}(p_{t})=-(1-p_{t})^{\gamma}\log(p_{t}).
\end{equation}

\noindent In the focal loss equation, $log(p_{t})$ represents cross-entropy, $(1-p_{t})^{\gamma}$ denotes the modulation factor and $\gamma$ is the focusing factor. The optimized parameters for focal loss are $\gamma=2$ and $(1-p_{t})=0.25$ \cite{focal_loss}.

\newpage

\subsection{Structuring multi-modal data fusion}

The different stages of fusion are demonstrated in Figure~\ref{fig:main_data_flow}. In the examples without fusion the satellite imagery is directly fed into the model (Figure~\ref{fig:main_data_flow}a). In early fusion, both the satellite imagery and GPS trajectory data are fused and then fed into the model (Figure~\ref{fig:main_data_flow}b). In the late fusion, both dataset are fed into in two separate models:

\begin{itemize}
    \item Type - 1: Deep learning model applied to satellite imagery but not applied to GPS trajectory (Figure~\ref{fig:main_data_flow}c).
    \item Type - 2: Deep learning model applied to both the satellite imagery and GPS trajectory (Figure~\ref{fig:main_data_flow}d).
\end{itemize}

After the late fusion networks, the two streams of data are combined into one using a fusion operation. Fusion methods involve matrix operations that combine multiple data sources into one. Table~\ref{table:fusion_operators} summarizes the fusion operations that are used in this study along with their respective equations.

\begin{table}[ht]
\caption{Fusion operators: A and B are input and C is the resulting tensor.}\label{table:fusion_operators}
    \centering
        \begin{tabular}{ccc}
            \toprule
            \textbf{Fusion Operator} & \textbf{Equation} & \textbf{Fusion Stage} \\ \midrule
            Concatenate & $C = [A\|B]$ & Early/Late \\
            Average & $C = (A + B)\;\circ\;0.5$ & Late \\
            Maximum & $C=max(A, B)$ & Late \\
            Multiply & $C = A\;\circ\;B$ & Late \\ 
            \botrule
        \end{tabular}
\end{table}

\subsection{Evaluation metric details}
\label{sec:method_metrics}
As recommended for segmentation tasks by \cite{metrics_reloaded}, the region-based metric IoU and the boundary-based metric Boundary-IoU \cite{boundary_iou} are adopted as the evaluation metric in this study.

The IoU of an individual example ($i$) is defined by the following equation \cite{cnn_accuracy1}:

\begin{equation}
    IoU_{i} = \frac{True\:Positives_{i}}{True\:Positives_{i} + False\:Positives_{i}+ False\:Negatives_{i}}.
\end{equation}

\noindent All values of this equation are in the number of pixels.

Boundary-IoU is a special form of the IoU metric. To calculate Boundary-IoU, the boundary pixels of the class are first extracted, and then the IoU metric is calculated using the same equation. Boundary IoU defined with the following equation: 


\begin{equation}
    Boundary\:IoU_{i} = \frac{(G_{d} \cap G) \cap (P_{d} \cap P)}{(G_{d} \cap G) \cup (P_{d} \cap P)},    
\end{equation}

\noindent where $G$ represents ground truth, $P$ represents prediction and $d$ represents the contour distance from mask pixels \cite{boundary_iou}.

The performance evaluation of the models are conducted using multiple test images. The mean value of IoU (mIoU) and Boundary-IoU (mBoundary-IoU) are considered as the final metric values and are calculated using the following equation \cite{hsgnet}:
\newline
\begin{equation}
    mIoU=\frac{1}{n}\sum_{n}^{i=1} {IoU_{i}}.
\end{equation}

\section{Experiments}
Experiments carried out in two different area for this study. This section provides the details about data, pre-processing steps, details of implementation of the methods explained in Section~\ref{sec:methodology}, the summary of the results and additional analysis.

\subsection{Data and pre-processing}
\begin{figure}[H]
  \subfigure[Input data: Istanbul test set]{\label{fig2:a}\includegraphics[width=0.5\linewidth]{ 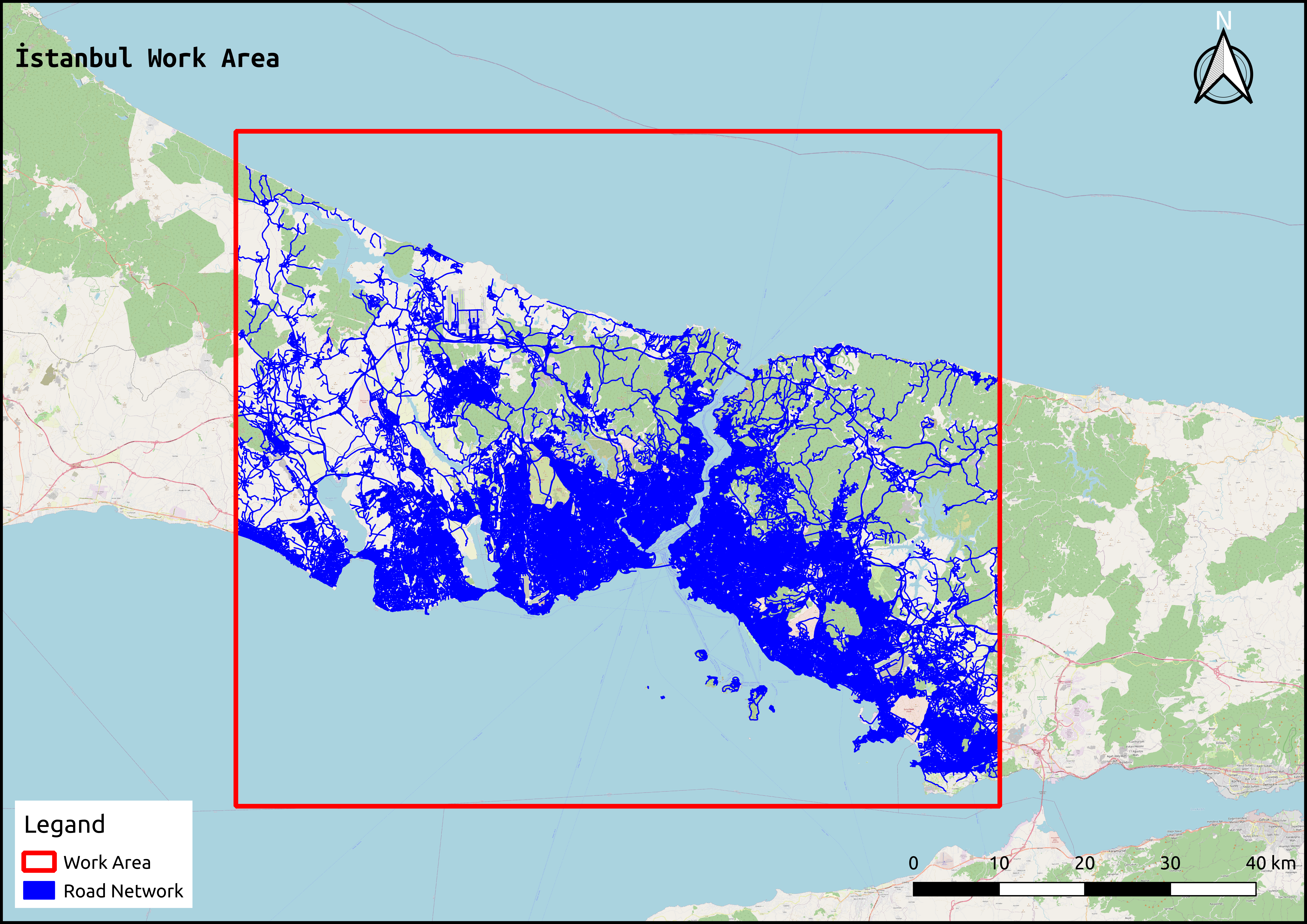}}
  \subfigure[Input data: Montreal test set]{\label{fig2:b}\includegraphics[width=0.5\linewidth]{ 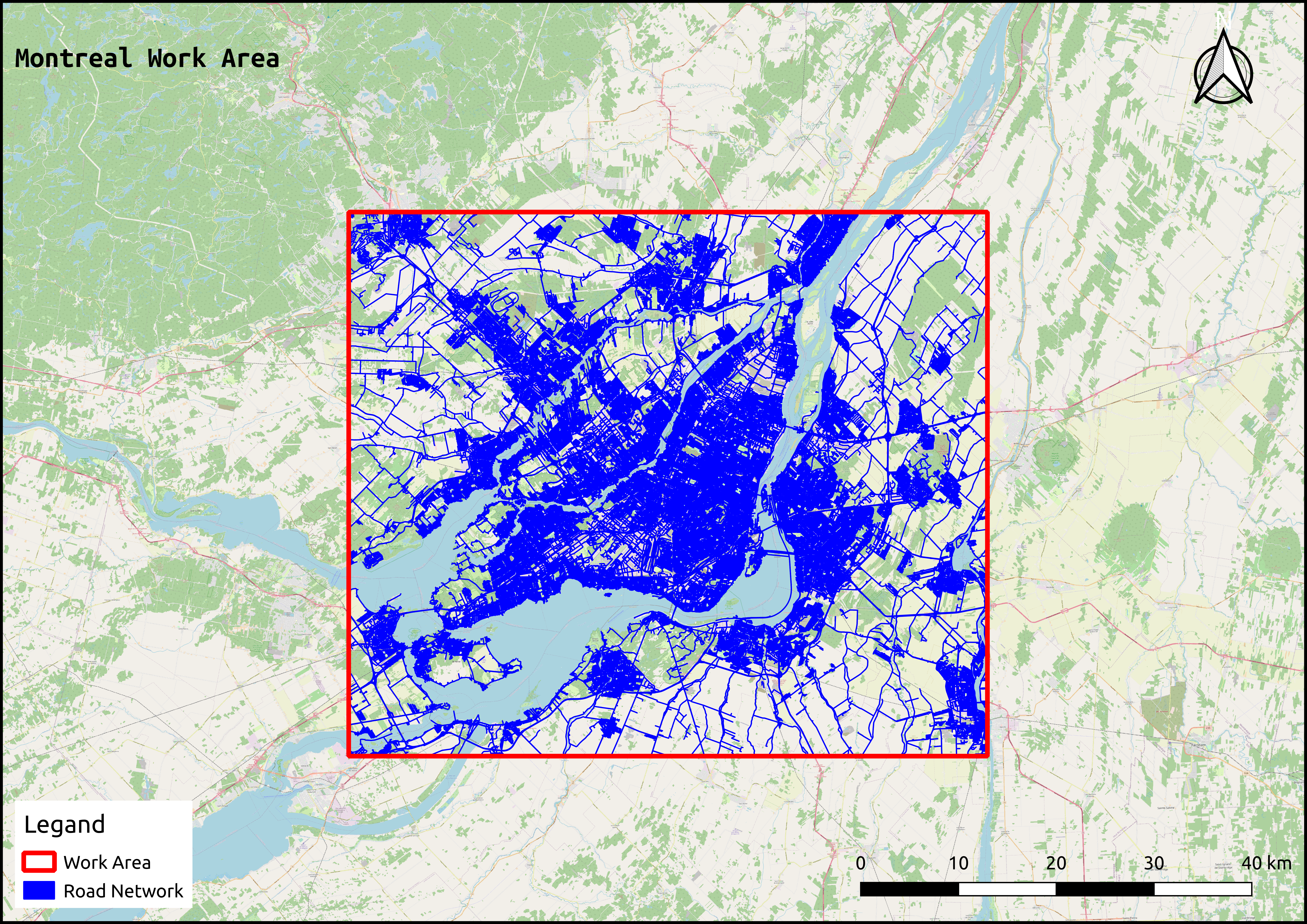}}
    \caption{Istanbul and Montreal work area.}
    \label{fig:work_area}
\end{figure}

Experiments were conducted in Istanbul - Turkey and Montreal - Canada. The work areas and corresponding road network coverage can be seen in Figure~\ref{fig:work_area}. These areas were chosen due to the availability of both GPS trajectory data and the Sentinel-2 data.

The GPS trajectory data in Istanbul was obtained from \cite{ibb_istac}. The data contains approximately 360 million GPS points from different months of 2020 and is collected from various types of vehicles such as cars and trucks. The data for Montreal was shared by \cite{mtl_trajet} and contains data from 2016 and 2017. The data consists of 40 million GPS points derived from passenger cars.

The satellite imagery used in experiments is derived from Sentinel-2 \cite{sentinel}. Sentinel-2 provides low-resolution ($10m/pixel$) multi-spectral satellite imagery, including red, green, blue (RGB) and infrared bands. The corresponding Sentinel-2 images taken around the same period as the GPS trajectory data were used in this study.

Since this study involves a supervised learning, a labeled data is required. Open Street Map (OSM) is an open map data source which is developed and maintained by volunteers \cite{osm}. The OSM data has been used in various road extraction studies \cite{osm_label1, Ayala2021_sentinel_2, Liu2022}. In this study, the label data was created using OSM data.

\begin{figure}[H]
\centering
  \includegraphics[width=0.75\linewidth]{ 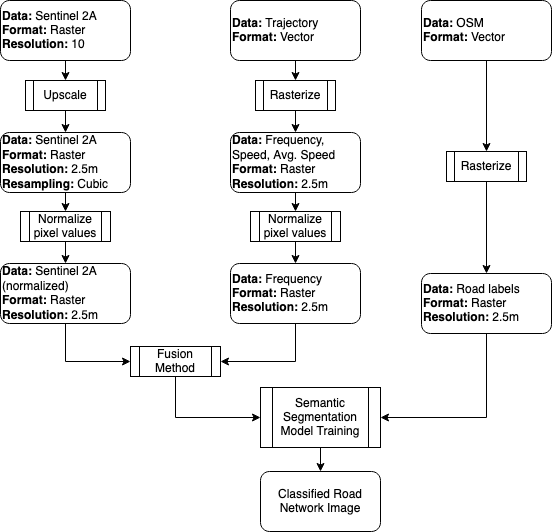}
    \caption{Data pre-processing details.}
    \label{fig:data_preprocessing}
\end{figure}

The data underwent pre-processing steps before training of the deep learning models. The details of the pre-processing steps are summarized in Figure~\ref{fig:data_preprocessing}. The RGB and infrared bands (RGB-I) were extracted from Sentinel-2 data and upscaled to $2.5m$ resolution using the cubic convolution re-sampling method, similar to \cite{Ayala2021_sentinel_2}. After upscaling, all bands were normalized to range of 0-1.

The GPS trajectory data was stored in tabular form and needed to be rasterized \cite{rasterization_necip}. To maintain the same resolution as the Sentinel-2 data, the GPS trajectory data was rasterized into $2.5m$ resolution imagery. The resultant imagery contains the frequency of GPS points per $2.5m$ x $2.5m$ square pixels.

The OSM data was stored in vector data format. To use it in this study, the OSM data was also rasterized. Since different classes of roads have different widths, a varying buffer was applied to the vector data, and rasterization was applied to the buffered data. The buffer values per road class are summarized in Table~\ref{table:osm_buffer}.

\begin{table}[ht]
    \caption{Road classes and applied buffer size.}\label{table:osm_buffer}
    \centering
      \begin{tabular}{cl}
            \toprule
            \textbf{Buffer (m)} & \textbf{OSM Road Class (fclass)}\\
            \midrule
            10 & "motorway", "primary", "secondary" \\
            6 & Remaining classes \\
            4 & \shortstack[l]{"footway" , "track", "service" , "steps" , "track\_grade1" , "track\_grade2" , \\"track\_grade3" , "track\_grade4" , "track\_grade5" , "track", "bridleway"} \\
            \botrule
        \end{tabular}
\end{table}

\newpage

All preprocessed data used in this study have been made available online to enable reproducibility and to be used as a benchmark in similar studies\footnote{The pre-processed data can be downloaded from following URL: \url{https://github.com/nagellette/sentinel_traj_nn/blob/master/Data.md}}

\subsection{Implementation and experimentation details}
This section provides information about the implementation of methods and additional details regarding the training of deep learning models that are used in the experiments.

The \textit{U-Net} \cite{unet}, \textit{ResUnet} \cite{resunet} and \textit{D-Linknet} \cite{d_linknet} models were implemented from scratch using the TensorFlow framework \cite{tensorflow}, and their respective architecture details were adopted from the corresponding publications. Additionally, the proposed to fusion stages were implemented using the same model architecture after adoption to the corresponding fusion model flow. The loss functions, MSE and BCE, were used as provided in TensorFlow framework. For focal loss, the implementation from TensorFlow Addons \cite{tensorflow_addons} was used with the default parameters specified in Section~\ref{sec:method_loss_functions}. The IoU metric was utilized as implemented in TensorFlow, while the Boundary-IoU metric \cite{boundary_iou} was implemented from scratch as it is explained in Section~\ref{sec:method_metrics}.

Both dataset were split into patches of size $512$ x $512$ pixels. To increase the dataset size, the raw patches were rotated at the angels of 45, 90, 135, 180, 225, 270 and 315 degrees with a 20\% overlap between neighboring patches. The total patch count for Istanbul and Montreal reached to 20,000 patches. The data was divided into train, validation, and test sets, with a ratio of 60\%, 20\% and 20\% ratio respectively.

All experiments were conducted using an NVIDIA Tesla V100 GPU with 16GB RAM. The models were trained with the Adam optimizer as it is implemented in TensorFlow, with a learning rate set to $0.001$. Training was performed using batches of patches which were randomly selected from the training set. The number of training epochs and batch sizes varied depending on the convergence of different models at different epochs and due to the memory limitations caused by the large model size for \textit{ResUnet} and \textit{D-Linknet} in late fusion experiments. Table~\ref{table:epochs_examples} summarizes the batch size, the number of epochs, and the number of batches per epoch used in the experiments. The training procedure was validated at the end of each epoch using $200$ randomly selected batches from the validation set. Once the training was completed, the performance of each model was evaluated using the IoU and Boundary-IoU metrics on $1000$ samples from the test set.

Cross work area training and testing were conducted to assess the generalization performance of the models on different dataset. For this purpose, the same model was trained using separately for Istanbul and Montreal, and Istanbul+Montreal together, and each of these trained models were evaluated on three test data combinations. For example, if a model was trained with data from Istanbul, it was evaluated using Istanbul, Montreal and Istanbul+Montreal test data where Istanbul+Montreal contains 50\% data from Istanbul test set and 50\% from Montreal test set.

\begin{table}[ht]
    \caption{The model training details: batch, epoch and number of batches per epochs in different experiments.}\label{table:epochs_examples}
    \centering
    \begin{tabular}{ccccc}
        \toprule
        \textbf{Model} & \textbf{Fusion Stage} & \textbf{Batch Size} & \textbf{Epochs} & \textbf{\# of batches/epoch} \\ \midrule
        U-Net & Early & 4 & 80 & 500 \\
        ResUnet & Early & 4 & 80 & 500 \\
        D-Linknet & Early & 4 & 150 & 500 \\
        U-Net & Late, Type-1 & 4 & 80 & 500 \\
        ResUnet & Late, Type-1 & 2 & 80 & 1000 \\
        D-Linknet & Late, Type-1 & 2 & 150 & 1000 \\
        U-Net & Late, Type-2 & 4 & 80 & 500 \\
        ResUnet & Late, Type-2 & 2 & 80 & 1000 \\
        D-Linknet & Late, Type-2 & 2 & 150 & 1000 \\ 
        \botrule
    \end{tabular}
\end{table}

All implementations used in this study have been made available online to enable reproducibility\footnote{The implementations of the methods and experiments can be downloaded from the following URL: \url{https://github.com/nagellette/sentinel_traj_nn}}.

\subsection{Results}

In the experiments, we considered Sentinel-2 only and early fusion as baseline results. The experiment results are summarized in the following tables: Table~\ref{table:baseline_results} shows the Sentinel-2 only and early fusion results, Table~\ref{fig:type_1_results} displays the Type-1 late fusion results, and Table~\ref{fig:type_2_results} presents the Type-2 late fusion results.

The best mIoU result with the Sentinel-2 only dataset was achieved by training \textit{ResUnet} on the Montreal dataset and evaluating it with the Montreal dataset using the BCE loss function (Table~\ref{table:baseline_results}). In early fusion experiments, \textit{ResUnet} achieved similar and slightly better mIoU results with the focal loss. However, all results showed a decrease in the mBoundary-IoU metric by a magnitude of $0.1{\sim}0.01$ compared to the mIoU score for the same experiment. Furthermore, there was a disagreement between the mIoU and mBoundary-IoU results when considering different loss functions in the same model and work area. For example, in the case of early fusion, in the Istanbul work area, the leading loss function was the focal loss with the mIoU metric, while it was MSE with the mBoundary-IoU metric. When considering cross work area evaluation, the results worsened when the training and evaluation work areas were different. 
The models trained and tested with the Montreal dataset achieved better results compared to the Istanbul and Istanbul+Montreal dataset. Although better results were achieved with models trained on the Montreal data, their mIoU performance dropped significantly (${\sim}0.2$) when compared to results of a dataset from another work area. This decrease was smaller for models trained on the Istanbul+Montreal dataset (${\sim}0.08$) and minimum for models trained on the Istanbul dataset (${\sim}0.04$).

The best mIoU and mBoundary-IoU results in Type-1 experiments were achieved with \textit{ResUnet} using the Montreal training dataset and testing it with the Montreal data using the MSE loss function and concatenation (0.767 in mIoU, 0.601 in mBoundary-IoU) (Table~\ref{fig:type_1_results}). These results showed a slight improvement compared to the early fusion experiments. Overall, \textit{ResUnet} was the leading model, and MSE was the leading loss function in the majority of the experiments when other variables were constant. The BCE loss function, when used with average and maximum fusion, caused a significant decrease in accuracy when other variables were constant. It is noteworthy that the multiply fusion method was on par with concatenation or even led in many experiments, especially when combined with focal loss. The disagreement observed between mIoU and mBoundary-IoU in Sentinel-2 only and early fusion experiments persisted. Additionally, the differences observed in the cross work area evaluation were still present, and these differences were increased in Type-1 experiments.

In Type-2 experiments, the best mIoU and mBoundary-IoU results were achieved with \textit{ResUnet} using the Montreal training dataset and testing it with the Montreal data using the MSE loss function and concatenation (0.784 in mIoU, 0.631 in mBoundary-IoU) (Table~\ref{fig:type_2_results}). This represents a significant improvement compared to Type-1 and early fusion experiments. Similar to Type-1, \textit{ResUnet} and MSE were the leading model and loss function, respectively. The decreased performance of BCE with average and maximum fusion methods still persisted, and the magnitude of accuracy decrease was greater compared to Type-1. Additionally, the observed disagreements between mIoU and mBoundary-IoU were still present, and the differences in cross-area evaluation were even more pronounced.

In the cross work area evaluation, both quantitative and qualitative (Figure~\ref{fig:cross_work_area_fig}) evaluations showed that the models' generalization was limited. However, the experimental results suggested that early fusion methods were able to generalize better compared to late fusion alternatives, although early fusion methods achieved lower mIoU and mBoundary-IoU scores. It was particularly significant that the models were able to generalize better to wider roads when GPS trajectory data was fused. Additionally, the models trained with Istanbul data performed worse in generalization compared to the Montreal data when other variables were constant. Moreover, the models trained and tested with Istanbul data achieved lower mIoU and mBoundary-IoU scores compared to the models trained and tested with Montreal data. The differences in land coverage propagation and settlement characteristics were considered the main reasons for this discrepancy. This aspect is further analyzed in Section~\ref{sec:texture_analysis} with a complexity similarity comparison, which examines the variable land coverage and settlement between the two cities.

\begin{sidewaystable}[ph!]
\caption{Baseline experiments with Sentinel RGB-I data.}
\label{table:baseline_results}
    \begin{tabular}{ccccccccccccccc}
        \toprule
        \multirow{3}{*}{\textbf{Work Area}} & \multirow{3}{*}{\textbf{Model}} & \multirow{3}{*}{\textbf{Test Area}} & \multicolumn{6}{c}{\textbf{Sentinel-2}} & \multicolumn{6}{c}{\textbf{Sentinel-2+GPS Trajectory}} \\ \cmidrule{4-15} 
         &  &  & \multicolumn{3}{c}{\textbf{mIoU}} & \multicolumn{3}{c}{\textbf{mBoundary IoU}} & \multicolumn{3}{c}{\textbf{mIoU}} & \multicolumn{3}{c}{\textbf{mBoundary IoU}} \\ \cmidrule{4-15} 
         &  &  & \textbf{BCE} & \textbf{Focal} & \textbf{MSE} & \textbf{BCE} & \textbf{Focal} & \textbf{MSE} & \textbf{BCE} & \textbf{Focal} & \textbf{MSE} & \textbf{BCE} & \textbf{Focal} & \textbf{MSE} \\ \hline
        \multirow{9}{*}{Istanbul} & \multirow{3}{*}{D-Linknet} & Istanbul & 0.641 & 0.643 & 0.649 & 0.520 & 0.510 & 0.524 & \textbf{0.659} & 0.663 & \textbf{0.662} & 0.533 & 0.522 & 0.534 \\
         &  & Ist.+Mont. & 0.603 & 0.616 & 0.607 & 0.514 & 0.506 & 0.515 & 0.643 & 0.656 & 0.639 & \textbf{0.528} & 0.521 & 0.527 \\
         &  & Montreal & 0.561 & 0.587 & 0.562 & 0.506 & 0.501 & 0.506 & 0.623 & 0.648 & 0.614 & 0.521 & 0.521 & 0.518 \\ \cmidrule{2-15} 
         & \multirow{3}{*}{ResUnet} & Istanbul & \textbf{0.657} & \textbf{0.648} & \textbf{0.659} & \textbf{0.531} & \textbf{0.517} & \textbf{0.535} & 0.644 & \textbf{0.664} & 0.661 & 0.527 & 0.523 & \textbf{0.536} \\
         &  & Ist.+Mont. & 0.593 & 0.610 & 0.595 & 0.515 & 0.510 & 0.517 & 0.619 & 0.655 & 0.627 & 0.519 & 0.521 & 0.524 \\
         &  & Montreal & 0.524 & 0.569 & 0.529 & 0.498 & 0.503 & 0.498 & 0.592 & 0.646 & 0.592 & 0.512 & 0.519 & 0.512 \\ \cmidrule{2-15} 
         & \multirow{3}{*}{U-Net} & Istanbul & 0.634 & 0.630 & 0.636 & 0.518 & 0.508 & 0.520 & 0.645 & 0.661 & 0.654 & 0.525 & 0.521 & 0.528 \\
         &  & Ist.+Mont. & 0.581 & 0.610 & 0.582 & 0.507 & 0.505 & 0.507 & 0.623 & 0.659 & 0.635 & 0.519 & 0.524 & 0.523 \\
         &  & Montreal & 0.526 & 0.590 & 0.524 & 0.495 & 0.502 & 0.493 & 0.599 & 0.658 & 0.615 & 0.513 & \textbf{0.526} & 0.518 \\ \hline
        \multirow{9}{*}{Ist.+Mont.} & \multirow{3}{*}{D-Linknet} & Istanbul & 0.618 & 0.630 & 0.622 & 0.513 & 0.507 & 0.514 & 0.648 & 0.651 & 0.645 & 0.524 & 0.514 & 0.520 \\
         &  & Ist.+Mont. & 0.650 & 0.649 & 0.648 & 0.526 & 0.514 & 0.527 & 0.677 & 0.675 & 0.673 & 0.539 & 0.526 & 0.533 \\
         &  & Montreal & 0.680 & 0.664 & 0.668 & 0.538 & 0.521 & 0.537 & \textbf{0.702} & 0.694 & 0.699 & 0.553 & 0.537 & 0.545 \\ \cmidrule{2-15} 
         & \multirow{3}{*}{ResUnet} & Istanbul & 0.614 & 0.637 & 0.633 & 0.515 & 0.511 & 0.519 & 0.646 & 0.633 & 0.640 & 0.527 & 0.508 & 0.523 \\
         &  & Ist.+Mont. & 0.656 & 0.669 & 0.671 & 0.533 & 0.522 & 0.539 & 0.671 & 0.664 & 0.668 & 0.541 & 0.520 & 0.539 \\
         &  & Montreal & \textbf{0.691} & \textbf{0.698} & \textbf{0.706} & \textbf{0.549} & \textbf{0.533} & \textbf{0.559} & 0.691 & 0.695 & 0.691 & \textbf{0.554} & 0.533 & 0.553 \\ \cmidrule{2-15} 
         & \multirow{3}{*}{U-Net} & Istanbul & 0.602 & 0.633 & 0.606 & 0.509 & 0.510 & 0.511 & 0.617 & 0.647 & 0.640 & 0.517 & 0.516 & 0.523 \\
         &  & Ist.+Mont. & 0.637 & 0.661 & 0.628 & 0.525 & 0.520 & 0.523 & 0.646 & 0.675 & 0.674 & 0.530 & 0.527 & 0.541 \\
         &  & Montreal & 0.664 & 0.685 & 0.642 & 0.540 & 0.530 & 0.532 & 0.671 & \textbf{0.703} & \textbf{0.703} & \textbf{0.543} & \textbf{0.540} & \textbf{0.558} \\ \hline
        \multirow{9}{*}{Montreal} & \multirow{3}{*}{D-Linknet} & Istanbul & 0.495 & 0.520 & 0.524 & 0.484 & 0.487 & 0.488 & 0.566 & 0.591 & 0.590 & 0.495 & 0.499 & 0.500 \\
         &  & Ist.+Mont. & 0.602 & 0.607 & 0.624 & 0.520 & 0.514 & 0.528 & 0.651 & 0.657 & 0.651 & 0.536 & 0.526 & 0.528 \\
         &  & Montreal & 0.703 & 0.690 & 0.722 & 0.555 & 0.540 & 0.566 & 0.730 & 0.724 & 0.705 & 0.575 & 0.553 & 0.554 \\ \cmidrule{2-15} 
         & \multirow{3}{*}{ResUnet} & Istanbul & 0.516 & 0.521 & 0.534 & 0.487 & 0.487 & 0.490 & 0.577 & 0.576 & 0.581 & 0.499 & 0.495 & 0.499 \\
         &  & Ist.+Mont. & 0.639 & 0.635 & 0.645 & 0.542 & 0.528 & 0.542 & 0.670 & 0.672 & 0.670 & 0.548 & 0.540 & 0.548 \\
         &  & Montreal & {\ul \textbf{0.760}} & \textbf{0.743} & \textbf{0.755} & \textbf{0.596} & \textbf{0.568} & \textbf{0.595} & \textbf{0.758} & {\ul \textbf{0.763}} & \textbf{0.757} & \textbf{0.596} & \textbf{0.583} & \textbf{0.597} \\ \cmidrule{2-15} 
         & \multirow{3}{*}{U-Net} & Istanbul & 0.505 & 0.517 & 0.486 & 0.486 & 0.487 & 0.480 & 0.598 & 0.602 & 0.591 & 0.502 & 0.500 & 0.499 \\
         &  & Ist.+Mont. & 0.627 & 0.626 & 0.597 & 0.535 & 0.521 & 0.523 & 0.675 & 0.674 & 0.670 & 0.543 & 0.531 & 0.543 \\
         &  & Montreal & 0.745 & 0.730 & 0.702 & 0.582 & 0.553 & 0.564 & 0.749 & 0.746 & 0.746 & 0.584 & 0.565 & 0.585 \\ 
        \botrule
    \end{tabular}
\end{sidewaystable}

\setcounter{figure}{4}

\begin{sidewaysfigure}
    \captionsetup{name=Table}
    \caption{Experiment results of Type-1 late fusion networks.}
    \centering
    \includegraphics[width=1\linewidth]{ 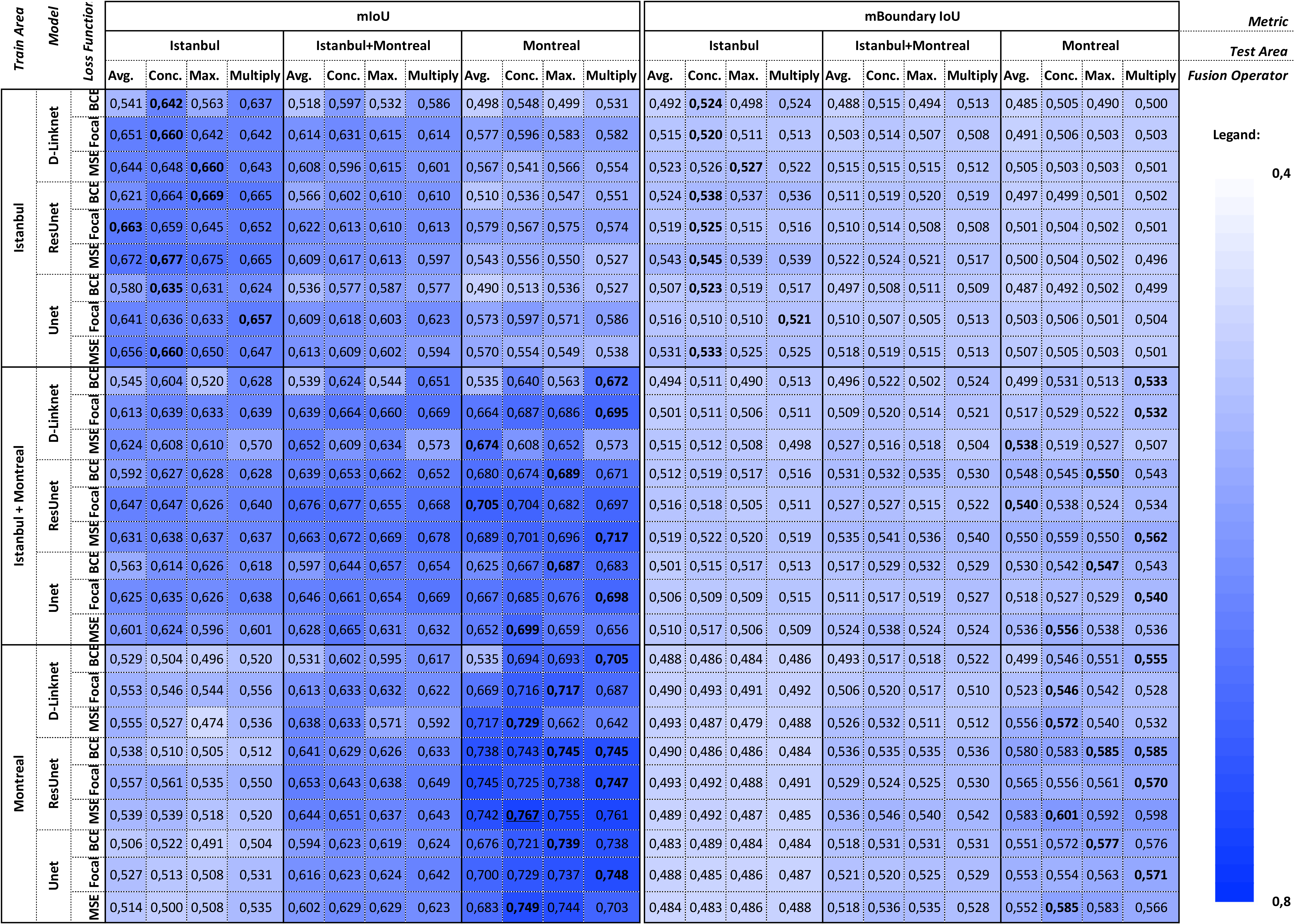}
     \label{fig:type_1_results}
\end{sidewaysfigure}

\begin{sidewaysfigure}
    \captionsetup{name=Table}
    \caption{Experiment results of Type-2 late fusion networks.}
    \centering
    \includegraphics[width=1\linewidth]{ 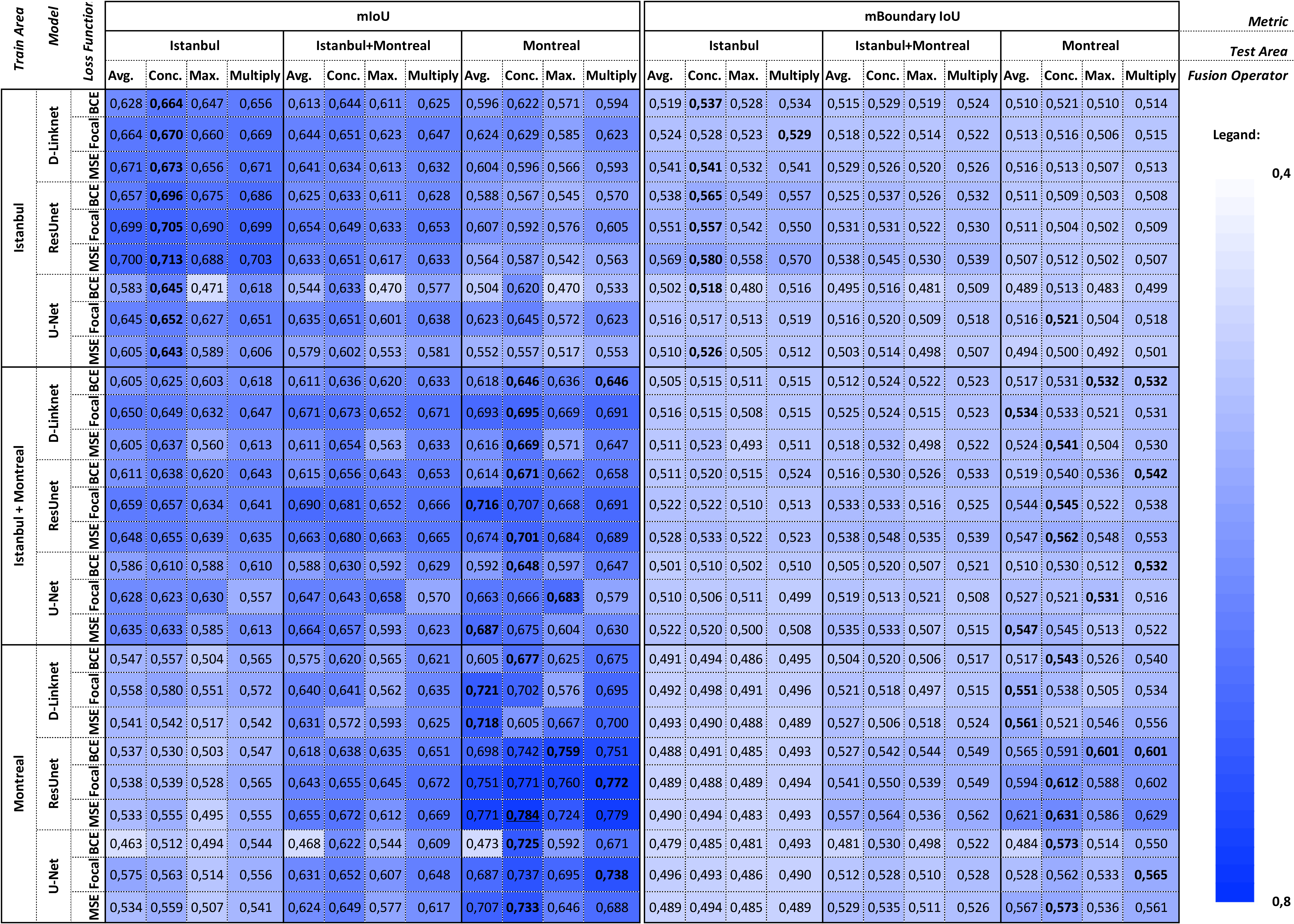}
    \label{fig:type_2_results}
\end{sidewaysfigure}

\setcounter{figure}{3}

\begin{figure}[H]
\begin{center}
      \subfigure[Input data: Istanbul test set]{\label{fig2:a}\includegraphics[width=0.47\linewidth]{ 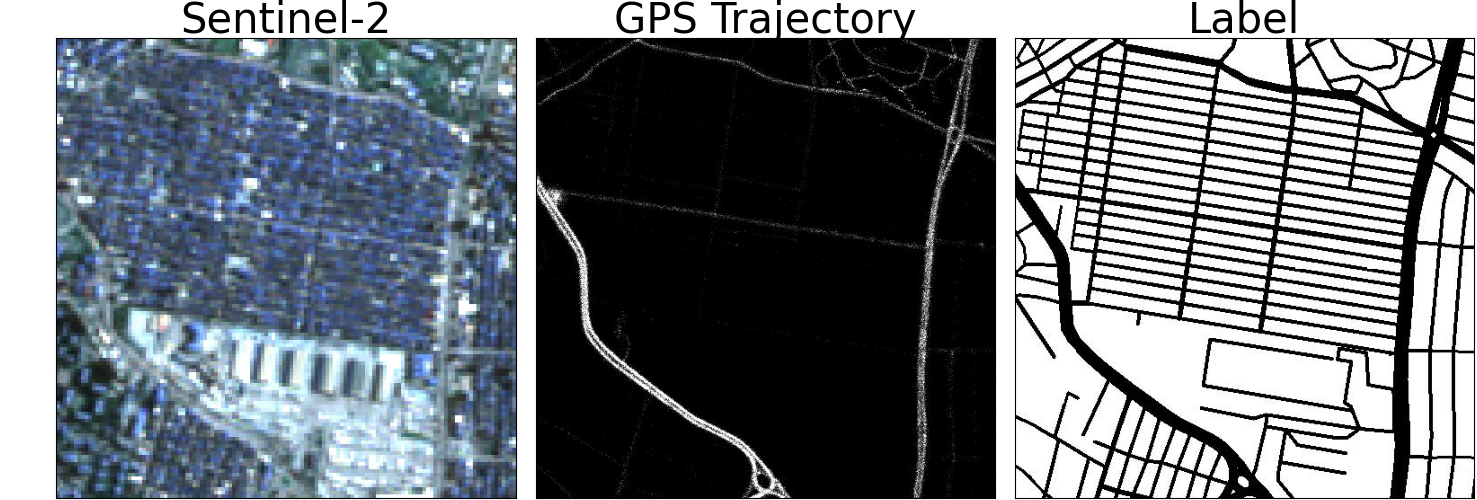}}
  \subfigure[Input data: Montreal test set]{\label{fig2:b}\includegraphics[width=0.47\linewidth]{ 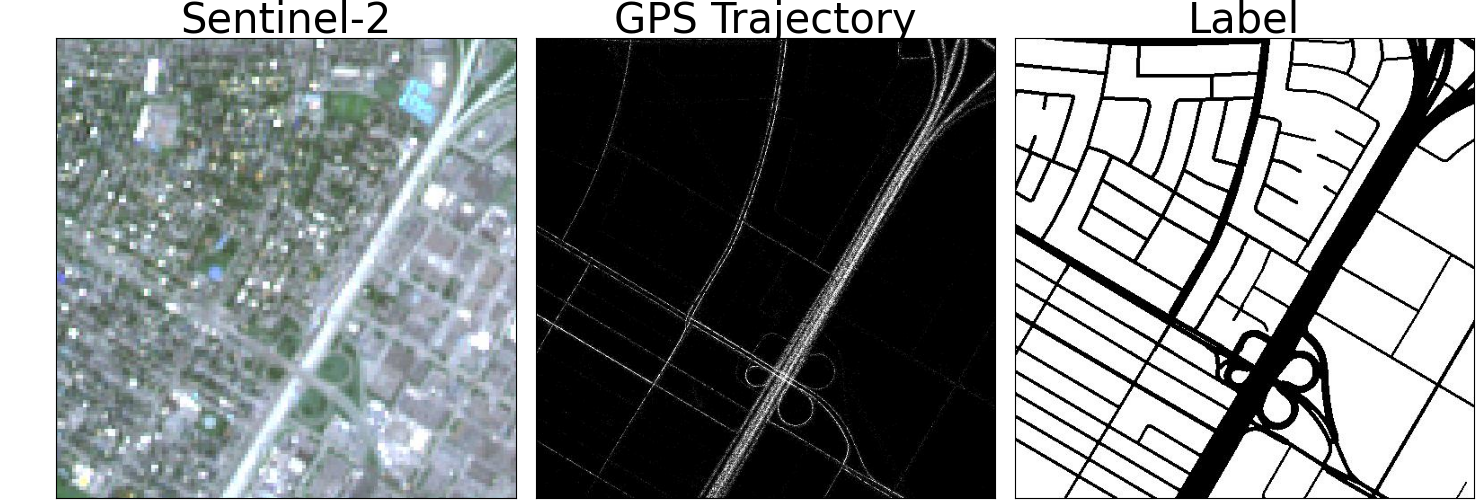}}\\
  \subfigure[Output data: Best performing models tested with Istanbul data.]{\label{fig2:c}\includegraphics[width=0.47\linewidth]{ 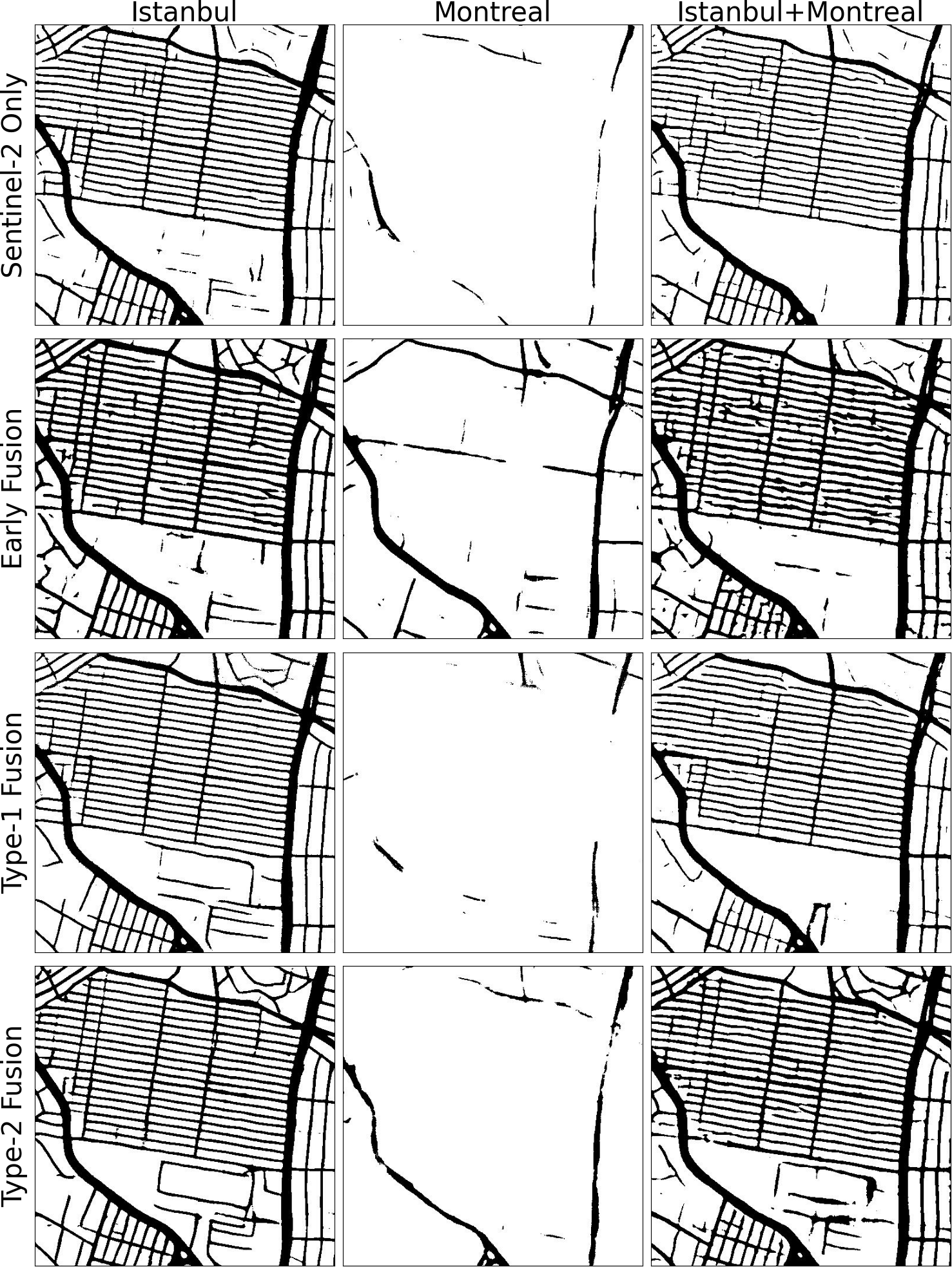}}
  \subfigure[Output data: Best performing models tested with Montreal data.]{\label{fig2:d}\includegraphics[width=0.47\linewidth]{ 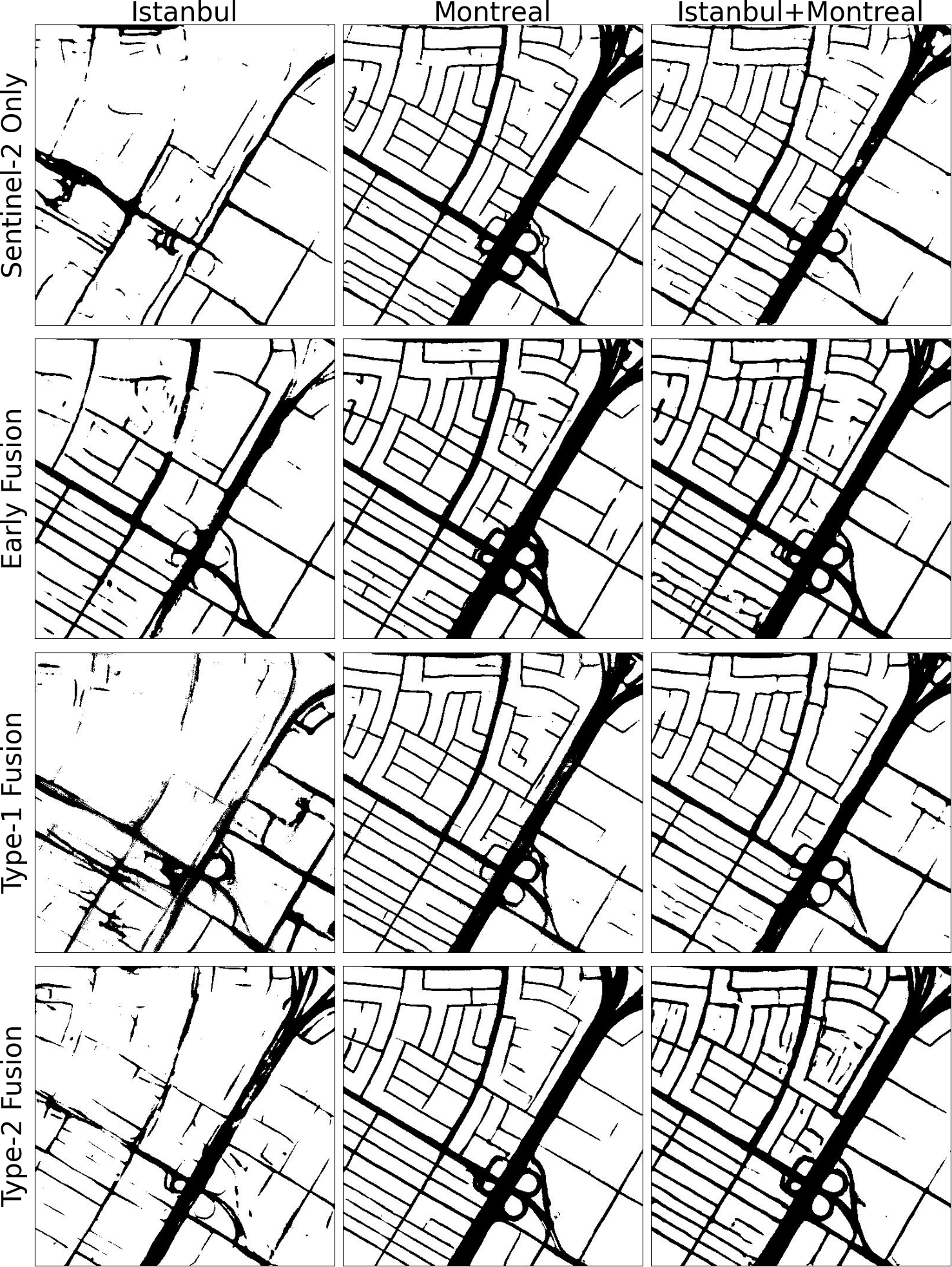}}
    \caption{Generalization capabilities of different model types which are trained in different dataset: (a) and (b) shows the input data from Istanbul and Montreal test set respectively. (c) and (d) provides the output of best performing models - the columns show the dataset which model is trained on, rows show the type of fusion in use.}
    \label{fig:cross_work_area_fig}
\end{center}
\end{figure}

Finally, Table~\ref{table:results_comp_literature} provides a summary of the best achieved results and their comparison to the results of \cite{Ayala2021_sentinel}, which served as the literature benchmark for road extraction using Sentinel-2 data. The best models trained with Istanbul and Montreal data surpassed the mIoU scores reported in the literature, particularly in the cases of late Type-1 and Type-2 networks, when GPS trajectory and Sentinel-2 data were utilized. In addition to IoU results mBoundary-IoU results are also available which can be used as the future benchmark for shape based comparison.

\setcounter{table}{6}

\begin{table}[h]
\caption{Comparison with benchmarks and our best performing models.}\label{table:results_comp_literature}
\centering
    \begin{tabular}{lcc}
    \toprule
    \textbf{Model} & \textbf{IoU} & \textbf{mBoundary IoU}\textsuperscript{*}\\ \midrule
    U-Net + Bicubic x4 \\ Overall \cite{Ayala2021_sentinel} & 0.6894 & - \\ \midrule
    U-Net + Bicubic x4 \\ Best \cite{Ayala2021_sentinel} & 0.7066 & - \\ \midrule
    ResUnet Type-2 \\ with Concatenation \& MSE Loss  \\trained in  Istanbul+Montreal \\ (Best in Istanbul test samples)\textsuperscript{*} & 0.713 & 0.580 \\ \midrule
    ResUnet \\ without GPS Trajectory fusion \\with BCE Loss trained in  Montreal\textsuperscript{*} & 0.760 & 0.596 \\ \midrule
    ResUnet Early fusion \\ with Focal Loss \\trained in Montreal\textsuperscript{*} & 0.763 & 0.583 \\ \midrule
    ResUnet Type-1 fusion \\ with Concatenation \& MSE Loss \\trained in Istanbul+Montreal\textsuperscript{*} & 0.767 & 0.601 \\ \midrule
    ResUnet Type-2 fusion \\ with Concatenation \& MSE Loss \\trained in Istanbul+Montreal\textsuperscript{*} & \textbf{0.784} & \textbf{0.631} \\ 
    \botrule
    \scriptsize{\textsuperscript{*} Our contributions} & &
    \end{tabular}
\end{table}

\subsection{Complexity analysis}
\label{sec:texture_analysis}

\begin{figure}[h]
  \subfigure[Entropy]{\label{fig2:a}\includegraphics[width=0.50\linewidth]{ 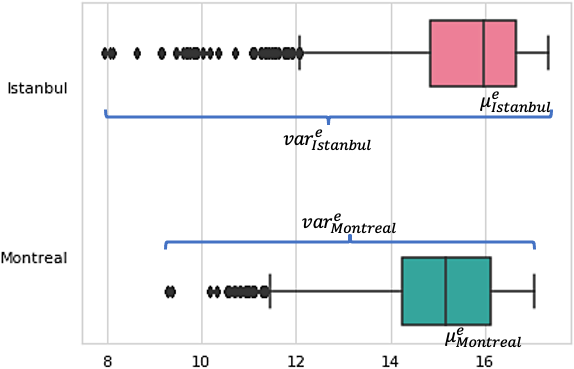}}
  \subfigure[Homogeneity]{\label{fig2:b}\includegraphics[width=0.50\linewidth]{ 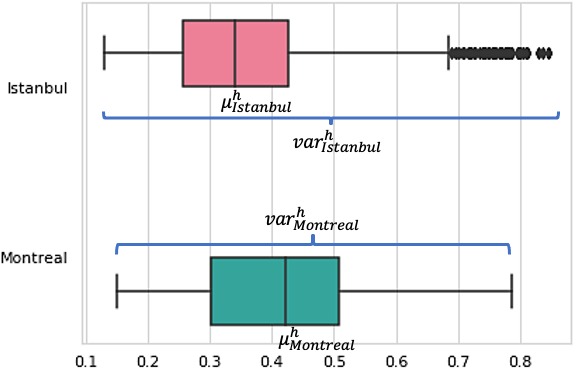}}
    \caption{Complexity assessment of training examples from different work areas: (a) entropy and (b) homogeneity distribution in Istanbul and Montreal.}
    \label{fig:complexity_analysis}
\end{figure}

Due to the differences in the metric results of the same models on different dataset, it is necessary to determine if the two dataset have similar inputs in terms of complexity. The evaluation results suggest that the complexity of the Istanbul dataset differs from that of the Montreal dataset, and the models trained on their respective work areas exhibit varying levels of accuracy. The complexity of an image dataset can be analyzed by measuring the differences in entropy \cite{entropy1} or the texture homogeneity derived from the Gray-Level Co-Occurrence Matrix (GLCM) \cite{glcm1, complexity_measure}. Entropy represents the uncertainty of a system \cite{entropy_giz}, while GLCM is a matrix that illustrates the spatial distribution of gray levels within an image, providing additional information such as texture, contrast, and correlation. In this study, entropy (calculated using \cite{entropy2} and homogeneity from GLCM (calculated using \cite{glcm2} are computed for each image patch from Istanbul and Montreal, and the distribution of these values are visualized in Figure~\ref{fig:complexity_analysis}.

The mean entropy value for Istanbul ($\mu^{e}_{Istanbul}$) is higher than that of Montreal ($\mu^{e}_{Montreal}$), indicating that, on average, the Istanbul examples exhibit higher levels of variability compared to the examples from Montreal ($var^{e}_{Istanbul}\:>\:var^{e}_{Montreal}$). Additionally, the variability of entropy examples in Istanbul is more diverse than in Montreal. On the other hand, in terms of homogeneity, the mean value for Montreal ($\mu^{h}_{Montreal}$) examples is higher than that of Istanbul ($\mu^{h}_{Istanbul}$). This implies that the examples in Montreal are more homogeneous and provide less complex information compared to those in Istanbul. Similar to entropy, the variability of examples in Istanbul is higher than the examples in Montreal ($var^{h}_{Istanbul}\:>\:var^{h}_{Montreal}$).

\section{Discussion and Concluding Remarks}
In this study, the performance of different deep learning models, loss functions, fusion approaches, and model generalization is evaluated for road extraction tasks using low-resolution satellite imagery and GPS trajectory data. The evaluation of the results is conducted using a region-based metric and a shape-based metric, with using a new benchmark dataset covering Istanbul and Montreal. The results indicate that \textit{ResUnet} outperforms \textit{U-Net} and \textit{D-Linknet} in road extraction tasks and achieves better results than the benchmark study by \cite{Ayala2021_sentinel} using low-resolution Sentinel-2 data.

Overall, the performance of road extraction results improves when GPS trajectory data is fused with satellite imagery, particularly in the case of late fusion Type-2 with concatenation and multiply methods. Among the evaluated loss functions, MSE performs the best, while focal loss and BCE perform slightly worse, with BCE demonstrating a significant drop in performance when used in combination with average and maximum fusion methods. Additionally, the evaluation metrics provide novel insights into road extraction. The shape-based mBoundary-IoU metric generally provides similar information to the region-based IoU metric, although there are instances of disagreement, indicating that IoU may not be always reliable considering the shape of the output.

Regarding model generalization, the consistency of results among different models suggests that early fusion performs better while cross work area when testing compared to Type-1 and Type-2 late fusion networks.

In addition to the above findings, an analysis is conducted to understand the performance differences when training on different work areas. This analysis evaluates the complexity of the Istanbul and Montreal datasets using entropy and homogeneity measures, and concludes that the Istanbul dataset is more complex compared to the Montreal dataset.

It is worth noting that in the field of semantic segmentation, there are more complex models available recently, including Transformer-based models. Additionally, other loss functions, regularization strategies, and fusion methods can be considered to further extend the findings of this study. Beyond the ablation studies reported in this paper, further exploration of such architectural engineering approaches is kept outside the scope of this paper. Moreover, the complexity analysis carried out in this study can be expanded with additional complexity measures and can be used as an additional factor for the models. 

\bmhead{Acknowledgement}
This preprint has not undergone peer review (when applicable) or any post-submission improvements or corrections. The Version of Record of this article is published in Earth Science Informatics [ESIN], and is available online at \url{https://doi.org/10.1007/s12145-023-01201-6}.

The numerical calculations reported in this paper were fully performed at TUBITAK ULAKBIM, High Performance and Grid Computing Center (TRUBA resources). Authors would like to thank Istanbul Metropolitan Municipality and City of Montreal for GPS trajectory dataset, European Space Agency (ESA) for Sentinel-2 data and OpenStreetMap Foundation and OpenStreetMap Contributers for OpenStreetMap data. This study is part of the Ph.D thesis conducted in Istanbul Technical University by the first author. Authors would like to thank the Ph.D thesis advancement monitoring committee members, Gulsen Kaya Taskin and Taskin Kavzoglu.

\bmhead{Code \& data availability statement}
The code of the experiments and the data used in the experiments of this study made available online and related website information shared within the article.


\bibliography{sn-article}

\end{document}